\documentclass[conference]{IEEEtran}
\IEEEoverridecommandlockouts
\usepackage{cite}

\usepackage{amsmath,amssymb,amsfonts}
\usepackage{algorithmic}
\usepackage[linesnumbered,algoruled,boxed,lined]{algorithm2e}
\usepackage{comment}

\usepackage[normalem]{ulem}

\usepackage{graphicx}
\usepackage{wrapfig}
\usepackage{caption}
\usepackage{subcaption}
\usepackage{textcomp}
\usepackage{xcolor}
\usepackage[inline]{enumitem}

\begin{document}

\title{Non-Contact Heart Rate Measurement from Deteriorated Videos  \\
\thanks{This research has been supported by the Academy of Finland 6G Flagship program under Grant 346208 and PROFI5 HiDyn under Grant 32629, and the InSecTT project, which is funded under the European ECSEL Joint Undertaking (JU) program under grant agreement No 876038.}
}

\author{
\IEEEauthorblockN{Nhi Nguyen$^{\star}$, Le Nguyen$^{\star}$, Constantino Álvarez Casado$^{\star}$, Olli Silvén$^{\star}$, Miguel Bordallo López$^{\star}$$^{\dagger}$}
\IEEEauthorblockA{
\textit{$^{\star}$University of Oulu} \\
\textit{$^{\dagger}$VTT Technical Research Centre of Finland}\\
Oulu, Finland \\
thinguy21@student.oulu.fi, le.nguyen@oulu.fi, constantino.alvarezcasado@oulu.fi, olli.silven@oulu.fi, miguel.bordallo@oulu.fi}
}    

\maketitle

\begin{abstract}

Remote photoplethysmography (rPPG) offers a state-of-the-art, non-contact methodology for estimating human pulse by analyzing facial videos. Despite its potential, rPPG methods can be susceptible to various artifacts, such as noise, occlusions, and other obstructions caused by sunglasses, masks, or even involuntary facial contact, such as individuals inadvertently touching their faces. In this study, we apply image processing transformations to intentionally degrade video quality, mimicking these challenging conditions, and subsequently evaluate the performance of both non-learning and learning-based rPPG methods on the deteriorated data. Our results reveal a significant decrease in accuracy in the presence of these artifacts, prompting us to propose the application of restoration techniques, such as denoising and inpainting, to improve heart-rate estimation outcomes. By addressing these challenging conditions and occlusion artifacts, our approach aims to make rPPG methods more robust and adaptable to real-world situations. To assess the effectiveness of our proposed methods, we undertake comprehensive experiments on three publicly available datasets, encompassing a wide range of scenarios and artifact types. Our findings underscore the potential to construct a robust rPPG system by employing an optimal combination of restoration algorithms and rPPG techniques. Moreover, our study contributes to the advancement of privacy-conscious rPPG methodologies, thereby bolstering the overall utility and impact of this innovative technology in the field of remote heart-rate estimation under realistic and diverse conditions.

\end{abstract}


\section{Introduction}
\label{sec:introduction}
Non-contact video-based measurement of human heart pulse, also known as remote photoplethysmography (rPPG), can be implemented with an ordinary camera, such as a webcam~\cite{poh2010}. This technique analyzes videos to capture the subtle change of skin color caused by cardiac pulses to measure the pulse rate. There are non-learning-based rPPG methods that are based on signal processing algorithms, such as color space transformation, temporal normalization, independent component analysis, and projection to a plane orthogonal to the skin tone~\cite{bocc2020, Casado2022}. These methods are efficient and do not require model training, making them well-suited for real-time applications.
More recently, deep learning methods have emerged as a powerful approach for accurately estimating heart rate by optimally learning features from video data. However, these methods often require a large amount of data for training the models and may not generalize well to unseen data from different settings.

Despite recent technological advancements, remote photoplethysmography (rPPG) remains a challenging task due to numerous factors that can affect skin color variation, such as changes in illumination, head movements, hardware limitations, and networking infrastructure. Moreover, the widespread use of face masks in response to the COVID-19 pandemic has posed a new challenge for existing rPPG technologies~\cite{Speth2022}. In addition to these technical challenges, there exist privacy concerns associated with the use of rPPG methods based on face images, which can be exploited by face recognition algorithms to identify individuals~\cite{Ali2021}.

In this article, we investigate the performance of nine non-learning and learning-based rPPG methods on three datasets with varying settings (UBFC-rPPG \cite{ubfc}, UCLA-rPPG \cite{ucla}, and UBFC-Phys \cite{phys}):
\begin{itemize}
    \item We apply image transformation algorithms to deteriorate the videos of facial regions. The transformation serves two objectives: challenging the rPPG methods and removing sensitive information about user identities.
    \item We show that these transformation methods affect the performance of rPPG methods. Especially, in order to understand the generalization of learning-based approaches, we evaluate pre-trained models on datasets which they have not been trained on.
    \item We evaluate three restoration techniques to improve the accuracy of vital sign measurement on deteriorated videos: an image-denoising technique~\cite{coll2005}, a numerical method~\cite{telea2004}, and a learning-based approach~\cite{nafnet}.
\end{itemize}

Based on our experiments, we can select the optimal combination of rPPG methods and restoration approaches on deteriorated videos. The results provide valuable insights into the robustness of non-learning and learning-based rPPG methods on challenging situations. In addition, this study lays the groundwork for developing privacy-aware rPPG techniques.


\section{Proposed Methodology}
We propose to evaluate the performance of rPPG methods under challenging conditions and investigate the effect of image degradation on their accuracy. For this, we employ a range of non-learning and learning-based remote photoplethysmographic (rPPG) approaches for extracting rPPG signals from facial videos. These approaches are elucidated in a subsequent section. To ensure consistency, we utilize a standardized pipeline for preprocessing the input videos before feeding them into the respective rPPG methods. This pipeline, based on two state-of-the-art frameworks  \cite{pyVHR2020}\cite{Casado2022}, consists of different rPPG transformation methods and includes other processing blocks such as face detection, facial landmark estimation, region of interest (ROI) selection, and resizing.

Our primary objective in this study is to assess the impact of image degradation on the accuracy of remote photoplethysmography (rPPG) methods. To achieve this, we employ a variety of image transformation techniques to simulate challenging conditions, generating deteriorated videos. By evaluating the performance of the rPPG methods under these conditions, we aim to gain insights into their robustness and adaptability in real-world scenarios. In order to mitigate the effects of degradation, we then apply image restoration techniques.

We seek to identify optimal combinations of rPPG methods and restoration techniques to improve the accuracy of heart-rate estimation in deteriorated videos. By examining the effectiveness of various restoration approaches, we aim to better understand the robustness and adaptability of the rPPG methods under challenging conditions, ultimately contributing to the development of more reliable remote heart-rate estimation systems.

\subsection{Preprocessing}
\label{sec:faceextraction}

In order to prepare the input videos for rPPG analysis, we employed a systematic preprocessing pipeline that consisted of several stages. First, we utilized a pre-trained face detection model called Multi-Task Cascaded Convolutional Network (MTCNN) \cite{facenet} to detect faces in each frame of the video. Subsequently, we applied an Ensemble of Regression Trees (ERT) model \cite{OneMillisecondFaceAlignment, alvarez2021} to identify 68 facial landmark points within the detected face region, as defined by the Multi-PIE landmark scheme \cite{Gross-multipie}.

\begin{figure}[h!]
  \begin{center}
    \includegraphics[width=\columnwidth]{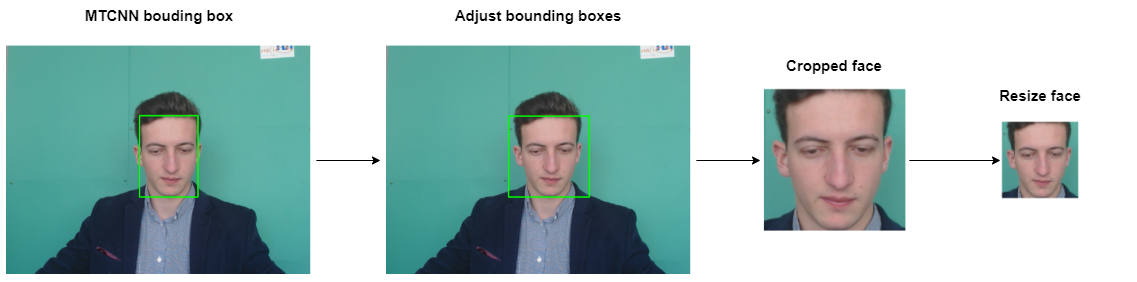}
  \end{center}
  \caption{Extraction of the the facial region used as input to the pipeline}
  \label{cropped_face}
\end{figure}

To further improve the precision of face detection and ROI selection, we developed an algorithm that adjusts the bounding boxes generated by MTCNN based on the detected landmark points. This algorithm takes the MTCNN bounding boxes and landmark points as input and returns adjusted square boxes centered on the face. The resulting square bounding boxes facilitate better performance in subsequent analysis steps. An example of this process is illustrated in Figure~\ref{cropped_face}.

Finally, the cropped face region was resized to dimensions $72 \times 72$ as the last step before applying the degradation and restoration techniques. By maintaining a consistent face region size across all videos, we ensured that the rPPG algorithms received standardized input data. This uniformity allows for a more accurate and reliable comparison of their performance under challenging conditions, as well as a fair evaluation of the effectiveness of various degradation and restoration methods.

\subsection{Image Deterioration Techniques}
\label{sec:transformation}

\begin{figure}
     \centering
     \begin{subfigure}[b]{0.18\columnwidth}
         \centering
         \includegraphics[width=\textwidth]{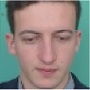}
         \caption{Original image}
         \label{fig:original}
     \end{subfigure}
     \begin{subfigure}[b]{0.18\columnwidth}
         \centering
         \includegraphics[width=\textwidth]{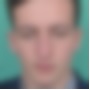}
         \caption{Motion blur}
         \label{fig:blur}
     \end{subfigure}
     \begin{subfigure}[b]{0.18\columnwidth}
         \centering
         \includegraphics[width=\textwidth]{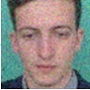}
         \caption{Gaussian noise}
         \label{fig:noise}
     \end{subfigure}
     \begin{subfigure}[b]{0.18\columnwidth}
         \centering
         \includegraphics[width=\textwidth]{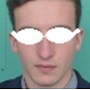}
         \caption{Eyemask}
         \label{fig:eyemask}
     \end{subfigure}
     \begin{subfigure}[b]{0.18\columnwidth}
         \centering
         \includegraphics[width=\textwidth]{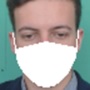}
         \caption{Facemask}
         \label{fig:facemask}
     \end{subfigure}
        \caption{Four image transformation techniques}
        \label{fig:imagetransformation}
\end{figure}

In order to simulate challenging conditions for remote photoplethysmography (rPPG) methods, we employ four image transformation techniques to degrade the videos. These techniques include motion blur, noise addition, occlusion of eyes, and incorporation of facemasks. We apply these transformations to the input facial regions.
The outcomes resulting from the application of these four image transformation techniques are illustrated in Figure \ref{fig:imagetransformation}.

\paragraph{Motion Blur} To emulate motion blur within the images, we employ a 2D Gaussian filter:
$
G(x, y)=\frac{1}{2 \pi \sigma^2} e^{-\frac{x^2+y^2}{2 \sigma^2}},
$
where $G(x, y)$ represents the Gaussian function at position $(x, y)$ in the image and $\sigma$ is the standard deviation of the Gaussian distribution, which governs the degree of blurring. The filter is convolved with the image using a kernel of size $k \times k$, with $k = 15$ in our experiments.

\paragraph{Adding Noise Artifacts} To simulate noise in images, we add Gaussian noise to the three color channels red (R), green (G), and blue (B), following these equations:
$I_R(x,y) = I_R(x,y) + N_R(x,y), I_G(x,y) = I_G(x,y) + N_G(x,y),$ and $I_B(x,y) = I_B(x,y) + N_B(x,y)$, respectively, where $I_R(x,y)$, $I_G(x,y)$, and $I_B(x,y)$ are the values of these channels at pixel (x,y) and $N_R(x,y)$, $N_G(x,y)$, and $N_B(x,y)$ are the amounts of Gaussian noise.

\paragraph{Occluding Eyes} In order to generate an occlusion in the form of sunglasses on a human face within an image, we compute the major and minor axes of the ellipse based on the distance between the eyes, the bridge of the nose, and the midpoint of the eyebrows, as depicted in Figure~\ref{fig:eyemask}.  We leverage facial landmark detection to accurately locate the key points necessary for constructing the mask, using the Dlib ERT \cite{dlib09}~ implementation and a state-of-the-art fast model~\cite{alvarez2021}. The resulting mask is then seamlessly integrated with the input image using a bitwise $\lor$ operation, resulting in final images featuring sunglasses-shaped masks.

\paragraph{Facemask} To generate a facemask for a facial image, we initially identify critical facial landmarks, such as the chin and nose tip, using the same approach as for eye occlusion. Subsequently, we create a polygon contour by employing a subset of the chin landmarks and the second point of the nose bridge. The interior of this contour is then filled with white to establish the mask. Ultimately, the mask is merged with the input image through a bitwise $\lor$ operation, resulting in the final image with the positioned facemask (see Figure~\ref{fig:facemask}).

\subsection{Image Restoration methods}

To mitigate the effects of image degradation, we employ various restoration techniques to reduce noise and remove facemasks from the facial images. To remove noise from the transformed datasets, we employ two approaches: a traditional denoising algorithm~\cite{coll2005} and a pre-trained deep learning model~\cite{nafnet} for image inpainting. However, when restoring facemasked effects, only the traditional method is used, as learning-based inpainting models trained on different data may result in unrealistic facial features. By employing these noise and facemask removal techniques, we aim to enhance the accuracy of the rPPG methods on degraded videos.

\paragraph{Non-local mean denoising}

The non-local means (NLM) algorithm \cite{coll2005} is a denoising technique that exploits the redundancy present in natural image statistics to remove noise from an image. Specifically, the algorithm assumes that similar patches in an image have similar noise characteristics, and thus averaging these patches can effectively reduce the noise while preserving the image structure. The NLM algorithm is effective in removing a variety of noise types, including Gaussian, impulse, and mixed noise from images. This approach has been shown to outperform other denoising techniques in terms of preserving image details and textures.

\begin{figure}
  \begin{center}
    \includegraphics[width=\columnwidth]{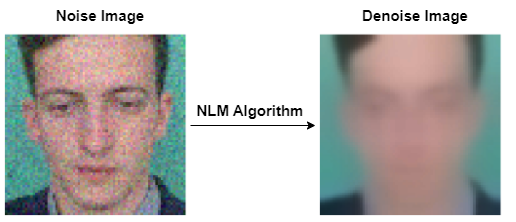}
  \end{center}
  \caption{Denoised image based on the NLM algorithm}
  \label{fig_nlm}
\end{figure}

The algorithm estimates a denoised value for a pixel $i$ in a discrete noisy image $v$, denoted as $NL[v](i)$. This value is obtained by taking a weighted average of all the pixels in the image, where the weights $w(i, j)$ depend on the similarity between pixels $i$ and $j$.  The similarity between two pixels $i$ and $j$ is based on the similarity of their intensity gray level vectors $v(\mathcal{N}_i)$ and $v(\mathcal{N}_j)$, where $\mathcal{N}_i$ and $\mathcal{N}_j$ are square neighborhoods centered at pixels $i$ and $j$, respectively.
Pixels that have similar grey-level neighborhoods to the pixel being denoised are given higher weights in the weighted average.
In our experiments, the NLM algorithm is implemented by utilizing the fastNlMeansDenoisingColored function in OpenCV~\cite{Bradski2000opencv}.
Figure~\ref{fig_nlm} shows an illustration of the NLM algorithm.

\paragraph{NAFNet Model Denoising}

\begin{figure}
  \begin{center}
    \includegraphics[width=\columnwidth]{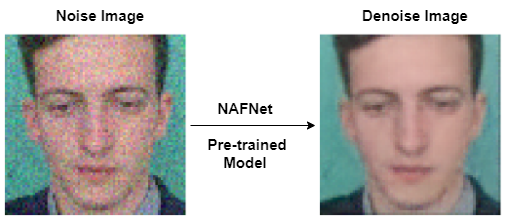}
  \end{center}
  \caption{Denoised image based on the NAFNet model}
  \label{fig_nafnet2}
\end{figure}

The NAFNet model~\cite{nafnet} is a deep learning network that has demonstrated exceptional performance on various multi-image restoration tasks, including denoising, deblurring, and super-resolution. This model was developed by customizing a U-net architecture~\cite{olaf2015} with skip connections, which involved removing nonlinear activation functions. A U-net typically consists of a contracting path and an expansive path.  For our study, we used a 32-layer NAFNet model, which was trained on the SIDD dataset \cite{abdel2018} using a batch size of 64 and a total of 400K training iterations. An example of the NAFNet model's performance is shown in Figure~\ref{fig_nafnet2}.

\paragraph{Fast marching method}

The fast marching method (FMM)~\cite{telea2004} is a partial differential equation-based approach that restores the missing region with information from the surrounding areas. The gray value at a point $p$ in the image can be approximated using a first-order approximation $I_q(p)$, given the image value $I(q)$ and its gradient $\nabla I(q)$ at a point $q$:

\begin{equation}\label{eq_ffm}
\begin{aligned}
&I_q(p) = I(q) + \nabla I(q)(p-q)\\
&I(p)=\frac{\sum_{q \in B_{\varepsilon}(p)} w(p, q)I_q(p)}{\sum_{q \in B_{\varepsilon}(p)} w(p, q)}
\end{aligned}
\end{equation}

\begin{figure}[h!]
  \begin{center}
    \includegraphics[width=\columnwidth]{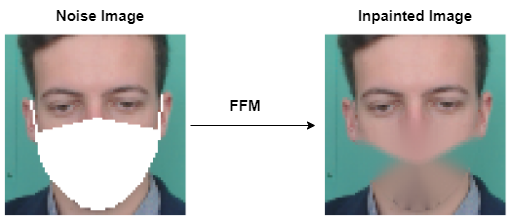}
  \end{center}
  \caption{Inpainted image using FFM to remove the facemask}
  \label{fig_telea2}
\end{figure}

In this work, the FMM was implemented with the INPAINT\_TELEA function in OpenCV~\cite{Bradski2000opencv}.  We visualize an example in Figure~\ref{fig_telea2}.

\section{Remote photoplethysmography methods}

Our experiments employ a total of six non-learning and three deep-learning rPPG methods. They are evaluated on both the original and the deteriorated videos.

\subsection{Non-learning rPPG methods}

Non-learning rPPG methods usually focus on the transformation of raw RGB signals into viable PPGs using a variety of methods\cite{bocc2020}. We employ six different transformations to assess the preprocessed data.

\paragraph{GREEN \cite{verk2008}} The GREEN method analyzes changes in the green color channel of the video frames, which are caused by variations in the blood volume on the human face during the cardiac cycle. The method computes the temporal average of the green channel values in the facial image. It uses a band-pass filter to isolate the heartbeat frequency range.

\paragraph{ICA \cite{poh2010}} This method utilizes Independent Component Analysis (ICA) to decompose the raw RGB traces and pick the second component, which typically contains a strong plethysmographic signal. Then, it applies the fast Fourier transform (FFT) on the selected signal and selects the highest power of the spectrum (within the operational range of $0.75 - 4$ Hz) as the pulse frequency.

\paragraph{CHROM \cite{haan2013}} This approach aims to eliminate the specular reflection component by using color difference or chrominance. It builds two orthogonal chrominance signals from the three color channels. The ratio of these is a candidate rPPG signal, which can be further improved with skin-tone standardization and a band-pass filter.

\paragraph{PBV \cite{haan2014}} The Pulse Blood Volume (PBV) vector is a linear combination of normalized color signals. The vector is used to obtain an orthogonal matrix to transform the RGB signals to the rPPG signals. The PBV vector depends on the light spectrum and camera sensor.

\paragraph{POS \cite{wang2017}} The Plane-Orthogonal-to-Skin method aims to remove specular reflections on the skin surface, similar to CHROM ~\cite{haan2013}, by defining a plane that is orthogonal to the skin tone in the normalized RGB color space. A tuning step is conducted to find the precise projection direction, where the specular and pulsatile components can be decomposed.

\paragraph{LGI \cite{pilz2018}} The Local Group Invariance (LGI) method maps skin pixel intensities into a lower dimensional space. It leverages the local invariance of the heart rate to create a robust feature space, which is less affected by interfering factors such as human movements and lighting conditions.

\subsection{Deep Learning rPPG Methods}

Deep learning based rPPG methods employ a large dataset of videos and reference signals to create a model. We employ three different pre-trained models, that have been fine-tuned for different datasets to improve their performance.

\paragraph{EfficientPhys~\cite{xiuliu2021}} The approach provides an end-to-end solution that does not require preprocessing steps such as face detection, segmentation, normalization, and color space transformation. Liu~\textit{et al.}~\cite{xiuliu2021} used a customized normalization module which consists of a different layer and a batch-normalization layer. The difference layer subtracts every two consecutive frames to reduce the global noise from lighting and motion. The batch-normalization layer has two learnable parameters for scaling and shifting to normalize
the difference frames within a training batch. After the normalization module, the network integrates a tensor-shift module and a hierarchical vision transformer architecture to learn the physiological signal from skin pixels. Liu~\textit{et al.}~\cite{liuxin2022} released an EfficientPhys model trained on the UBFC-rPPG dataset~\cite{ubfc}.

\paragraph{ContrastPhys~\cite{sun2022}} This approach leverages a 3DCNN model to extract multiple rPPG signals from each video at various spatiotemporal locations.
Then, the model is trained with a contrastive loss function so that the mean squared error between rPPG sequences from the same video is small while that of rPPG signals from different videos is large. The ContrastPhys~\cite{sun2022} model used in the experiments was trained on the UBFC-rPPG dataset~\cite{ubfc}.

\paragraph{PhysFormer~\cite{yu2021physformer}} The architecture is based on the temporal difference transformer. An RGB video is passed through convolution, batch normalization, ReLU, and MaxPool layers. The output is fed into the temporal difference transformer blocks to generate the global-local rPPG features, which are then projected to the 1D rPGG signal. The training strategy was based on the curriculum learning paradigm. The loss function was formulated as a combination of negative Pearson loss, frequency cross-entropy loss, and label distribution loss.
In this study, the PhysFormer model was trained on the VIPL-HR dataset~\cite{vipl}.

In our study, we selected sample face clips with dimensions of 320x72x72 ($T \times H \times W$), as this size is suitable for all three pre-trained models: In this context, the EfficientPhys model was specifically trained on samples of this size, making it appropriate for our purposes. The ContrastPhys model can accommodate any sample size, as its structure is not dependent on the input sample's size, allowing us to directly use the same size. In contrast, the PhysFormer model, was originally trained using an input size of 160×128×128. After passing through the header modules, the model produces a final feature of size 96x40x4x4, where 96 represents the depth of the features ($D$), and 40 represents $\frac{T}{4}$. By using an input sample size of 320x72x72, we can maintain the same performance of the previous modules while altering only the size of the final feature to 96x80x2x2. This sample size does not impact the performance of the previous modules in any way. Finally, passing the final feature through the predictor module of the PhysFormer model will produce the rPPG signal.

\section{Experiments and Analysis}

\subsection{Datasets and Protocol}
Three datasets used in our experiments are: UBFC-rPPG \cite{ubfc}, UCLA-rPPG \cite{ucla}, and UBFC-Phys \cite{phys}. Their properties are summarized in Table~\ref{tab:rppg_data}.

\captionsetup[table]{justification=raggedright,singlelinecheck=off}
\begin{table}
\footnotesize
\setlength{\tabcolsep}{1.3em}
\def\arraystretch{1.1}%
\begin{center}
  \caption{rPPG datasets in the experiments are UBFC-rPPG \cite{ubfc}, UCLA-rPPG \cite{ucla}, and UBFC-Phys~\cite{phys}}.
  \label{tab:rppg_data}
\begin{tabular}{c|c|c|c|}
\cline{2-4}
                                & Subjects & Video per subject & Videos                              \\ \hline
\multicolumn{1}{|c|}{UBFC-rPPG} & 50       & 1 video           & 50                           \\ \hline
\multicolumn{1}{|c|}{UCLA-rPPG} & 98       & 4-5 videos        & 488  \\ \hline
\multicolumn{1}{|c|}{UBFC-Phys} & 56       & 3 videos          & 168                          \\ \hline
\end{tabular}
\end{center}
\end{table}

UBFC-rPPG \cite{ubfc} contains 50 videos that were synchronized with a pulse oximeter finger-clip sensor for collecting ground truth data. Each video has a length of approximately 2 minutes and was recorded at a frame rate of 28-30Hz, with a resolution of 640x480 in uncompressed 8-bit RGB format. The first subset, UBFC1-rPPG, consists of eight videos in which participants were instructed to remain stationary with a signal sample rate of 62. The second subset, UBFC2-rPPG, includes 42 videos with a signal sample rate of 29-30 in which participants played a time-sensitive mathematical game designed to increase their heart rate while also simulating a realistic human-computer interaction scenario.

UBFC-Phys \cite{phys} consists of videos of 56 subjects. The participants were recorded with an Edmund Optics EO-23121C RGB digital camera, which used MotionJPEG compression and a frame rate of 35 frames per second, and a frame resolution of 1024 x 1024 pixels. Ground truth data were collected with an Empatica E4 wristband, including Blood Volume Pulse (BVP), skin temperature, and Electrodermal Activity (EDA) responses. During the experiment, participants experienced social stress through a three-step process involving a resting task (UBFC-Phys T1), a speech task (UBFC-Phys T2), and an arithmetic task (UBFC-Phys T3).

UCLA-rPPG \cite{ucla} includes 98 subjects. For each subject, 4-5 videos were recorded, each lasting about 1 minute and consisting of 1790 frames at 30fps. After removing any erroneous videos, a total of 488 videos were included in the dataset. All videos in the dataset were uncompressed and synchronized with ground truth heart rate data. To reduce redundancy in the dataset, only one video per subject was selected for our analysis. Specifically, the second video from each subject was used, resulting in a total of 98 videos.

In order to ensure uniformity in video duration and a fair evaluation, the videos were trimmed and downsampled. We extracted a 16-second segment from the middle of each video to avoid unusual actions. Subsequently, the selected segment was downsampled to 20 frames per second.
Furthermore, to enable consistent analysis across all datasets, the rPPG signal was also downsampled to match the number of video frames. Since the rPPG signal is periodic in nature, we utilized a Fourier-based method for resampling to preserve the information content of the signal and minimize the introduction of artifacts. Finally, after resampling, the data were standardized to remove any remaining noise or outliers that could impact the accuracy of our results. By implementing these techniques, we were able to ensure a uniform and reliable evaluation of the video data. The 16 seconds estimated rPPG waveform obtained from the rPPG method is filtered by applying a 2nd-order Butterworth filter with cut-off frequencies of 0.75 and 2.5 Hz to achieve the predicted HR.
To measure the quality of the estimated HR (in BPM), three common metrics are used: mean absolute error (MAE), root mean square error (RMSE) and mean absolute percentage error (MAPE).

\subsection{Evaluation of deterioration and restoration}

We evaluate the effectiveness of image restoration techniques to enhance the deteriorated videos for rPPG methods. The quality of images before and after restoration is assessed using Peak Signal-to-Noise Ratio (PSNR) and Structural Similarity Index (SSIM). Table \ref{tab:r_data1} illustrates that the NAFNet pre-trained denoising model and FFM inpainting method lead to a significant enhancement in image quality, while the NLM method does not provide much improvement.
The use of the FFM for inpainting on the facemask data has led to a slight improvement in image quality compared to the original, as showed in Table \ref{tab:r_data1}. 

\captionsetup[table]{justification=raggedright,singlelinecheck=off}
\def\arraystretch{1.2}%
\setlength{\tabcolsep}{1.2em}
\begin{table*}[ht!]
\begin{center}
  \caption{Perceived quality metrics of deteriorated and restored images compared to original images}
  \label{tab:r_data1}
\begin{tabular}{c|cc|cc|cc|cc|cc|} 
\cline{2-11}
                     & \multicolumn{2}{c|}{UBFC-rPPG}                   & \multicolumn{2}{c|}{UCLA-rPPG}        & \multicolumn{2}{c|}{UBFC-Phys T1}                & \multicolumn{2}{c|}{UBFC-Phys T2}                & \multicolumn{2}{c|}{UBFC-Phys T3}         \\ 
\cline{2-11}
                     & \multicolumn{1}{c|}{PSNR} & SSIM                 & PSNR           & SSIM                 & \multicolumn{1}{c|}{PSNR} & SSIM                 & \multicolumn{1}{c|}{PSNR} & SSIM                 & \multicolumn{1}{c|}{PSNR} & \multicolumn{1}{c|}{SSIM}           \\ 
\hline
\multicolumn{1}{c}{} & \multicolumn{10}{c}{\textcolor{orange}{Gaussian noise \& Denoised data}}                                                                                                                                                                           \\ 
\hline
Noise data                & 29.06                     & 0.64                 & 29.17          & 0.67                 & 29.07                     & 0.64                 & 29.07                     & 0.65                 & 29.07                     & 0.65           \\
NLM                  & 29.32                     & 0.63                 & 29.79          & 0.66                 & 29.7                      & 0.66                 & 29.64                     & 0.66                 & 29.67                     & 0.66           \\
NAFNet               & \textbf{31.62}            & \textbf{0.89}        & \textbf{30.71} & \textbf{0.84}        & \textbf{31.29}            & \textbf{0.87}        & \textbf{31.12}            & \textbf{0.87}        & \textbf{31.17}            & \textbf{0.87}  \\ 
\hline
\multicolumn{1}{c}{} & \multicolumn{10}{c}{\textcolor{purple}{Facemask \& Inpainting data}}                                                                                                                                                                              \\ 
\hline
Facemask                   & 33.58                     & 0.63                 & 33.33          & 0.61                 & 33.4                      & 0.64                 & 33.49                     & 0.66                 & 33.52                     & 0.65           \\
FFM                & \textbf{34.24}            & \textbf{0.78}        & \textbf{33.99} & \textbf{0.76}        & \textbf{33.89}            & \textbf{0.78}        & \textbf{33.8}             & \textbf{0.76}        & \textbf{33.79}            & \textbf{0.76}  \\ 

\hline
\end{tabular}
\end{center}
 \vspace{1mm}
\footnotesize{\textbf{Note:} The bold number represents the best result of each transformation.}

\end{table*}

\subsection{rPPG Evaluation on original datasets }

The results, as shown in Table \ref{tab:p_data1}, reveal that the EfficientPhys pre-trained model surpasses other methods in terms of MAE, RMSE, MAPE on  UBFC-rPPG, UCLA-rPPG, and UBFC-Phys T1 datasets, and achieved comparable results on UBFC-Phys T2 and UBFC-Phys T3 datasets.
Among the non-learning methods, POS appears to be the best performer, with the lowest error shown in all three metrics on UBFC-rPPG and UCLA-rPPG datasets. However, CHROM methods slightly outperformed POS on UBFC-Phys T1 datasets. It is noteworthy that the results on all methods for UBFC-Phys T2 and T3 datasets, which correspond to speech and arithmetic actions, are not very satisfactory, while the PhysFormer pre-trained model produced the lowest error.
Hence, we further evaluate EfficentPhys and POS on the transformed datasets including deteriorated videos.

\definecolor{orange}{HTML}{C2570E}
\definecolor{bluet}{HTML}{294483}
\definecolor{green}{HTML}{298323}
\definecolor{red}{HTML}{A31313}
\setlength{\tabcolsep}{0.35em}
\captionsetup[table]{justification=raggedright,singlelinecheck=off}
\def\arraystretch{1.3}
\begin{table*}[ht!]

\begin{center}
  \caption{Heart rate estimation error on UBFC-rPPG~\cite{ubfc}, UCLA-rPPG~\cite{ucla} and UBFC-Phys~\cite{phys}}
  \label{tab:p_data1}
\begin{tabular}{c|ccc|ccc|ccc|ccc|ccc|}
    
    
    \cline{2-16}
      \multicolumn{1}{c|}{}
    & \multicolumn{3}{c|}{UBFC-rPPG~\cite{ubfc}}
    & \multicolumn{3}{c|}{UCLA-rPPG~\cite{ucla}} 
    & \multicolumn{3}{c|}{UBFC-Phys T1}
    & \multicolumn{3}{c|}{UBFC-Phys T2}
    & \multicolumn{3}{c|}{UBFC-Phys T3}
    \\                                             
    
    \hline         
       \multicolumn{1}{c|}{\textcolor{green}{Non-learning methods}} 
    & \multicolumn{1}{c|}{MAE} 
    & \multicolumn{1}{c|}{RMSE} 
    & \multicolumn{1}{c|}{MAPE}
    & \multicolumn{1}{c|}{MAE} 
    & \multicolumn{1}{c|}{RMSE} 
    & \multicolumn{1}{c|}{MAPE} 
    & \multicolumn{1}{c|}{MAE} 
    & \multicolumn{1}{c|}{RMSE} 
    & \multicolumn{1}{c|}{MAPE}
    & \multicolumn{1}{c|}{MAE} 
    & \multicolumn{1}{c|}{RMSE} 
    & \multicolumn{1}{c|}{MAPE}
    & \multicolumn{1}{c|}{MAE} 
    & \multicolumn{1}{c|}{RMSE} 
    & \multicolumn{1}{c|}{MAPE}
    \\ 

\hline
GREEN \cite{verk2008} & 29.77 & 40.08 & 27.64 & 8.99 & 15.70 & 10.98 & 16.95 & 25.04 & 19.45 & 18.5 & 22.79 & 23.75 & 19.63 & 23.74 & 24.11 \\
LGI \cite{pilz2018} & 27.75 & 38.85 & 26.05 & 4.26 & 11.15 & 5.04  & 7.99 & 15.81 & 9.68 & 17.2 & 21.89 & 23.31 & \uline{14.86} & \uline{20.53} & \textbf{18.32} \\
PBV  \cite{haan2014} & 26.16 & 38.48 & 24.25 & 7.94 & 14.32 & 9.97 & 12.39 & 21.53 & 14.63 & 16.78 & 20.97 & 22.03  & 20.13 & 25.41 & 25.55 \\
ICA \cite{poh2010} & 21.14 & 32.29 & 19.35 & 4.57 & 10.73 & 5.45 & 10.09 & 17.99 & 12.62 & 16.62 & 21.19 & 22.63 & 18.58 & 22.96 & 23.15 \\
CHROM \cite{haan2013} & 10.12 & 21.56 & 9.04 & 2.15 & 5.63 & 2.72 & \uline{6.03} & \uline{13.06} & \uline{8.45} 
                      & \textbf{15.03} & \textbf{19.76} & \textbf{22.57} & 14.94 & 21.17 & 21.88 \\
POS \cite{wang2017} & \uline{9.75} & \uline{19.27} & \uline{8.88} & \uline{1.46} & \uline{4.46} & \uline{1.93} 
                    & \uline{6.03} & 13.93 & 8.52& 15.57 & 20.55 & 22.64 & \textbf{14.65} & \textbf{20.42} & 21.97 \\

\hline       
      \multicolumn{1}{c|}{\textcolor{orange}{Deep learning methods}}           
    & \multicolumn{3}{c|}{} 
    & \multicolumn{3}{c|}{} 
    & \multicolumn{3}{c|}{} 
    & \multicolumn{3}{c|}{} 
    & \multicolumn{3}{c|}{} \\
\hline

EfficientPhys \cite{xiuliu2021} & \textbf{5.11} & \textbf{13.05} & \textbf{4.75} & \textbf{0.57} & {\textbf{1.74}} & \textbf{0.81} &                                      \textbf{4.35} & \textbf{9.46} & \textbf{6.65} & \uline{15.32} & \uline{19.91} & \uline{23.99} & 15.65 & 20.38 & 23.04  \\
ContrastPhys \cite{sun2022} & 26.02 & 37.79 & 23.25 & 5.43 & 12.12 & 6.30 & 11.55 & 19.70 & 13.76 & 15.65 & 20.03 & 20.71 & 16.41 & 21.08 & \uline{20.74} \\
PhysFormer \cite{tete2021} & 23.81 & 27.96 & 23.11 & 9.23 & 11.53 & 13.00 & 12.89 & 15.98 & 16.06 & 11.89 & 14.26 & 17.77 & 12.85 & 15.76 & 17.86 \\

\hline
\end{tabular}
\end{center}
\vspace{1mm}
\footnotesize{\textbf{Note:} The bold number represents the best result, while the underlined number represents the second-best.}
\vspace{-5mm}
\end{table*}


\subsection{rPPG Evaluation on Deteriorated Datasets}

Evaluating the EfficentPhys pre-trained model and the POS non-learning method on the transformed datasets helps us to understand their feasibility in challenging scenarios.
The original datasets were deteriorated by adding motion blur, noise artifacts, occluded eyes, and facemasks to deteriorate videos in UBFC-rPPG, UCLA-rPPG, and UBFC-Phys, as described in Section~\ref{sec:transformation}.

Based on the comparison in Tables \ref{tab:p_data2}, we conclude that the EfficientPhys pre-trained model delivers the best results for UBFC-rPPG, UCLA-rPPG, and UBFC-Phys T1 datasets, regardless of whether the data is blurred or occluded eyes are present.
However, for noisy and face mask videos, both EfficientPhys and POS methods are affected, but the POS method performs relatively better than EfficientPhys, except for the UBFC-rPPG dataset with face masks.
Regarding the UBFC-Phys T2 and T3 datasets, the results of the POS method showed a slight improvement when face masks were applied to the original videos. This suggests that when there is significant head movement in the images, applying face masks can further enhance the performance of the POS method. Despite this, the improvement is not significant, and the results remain relatively poor for datasets with substantial head movements.

\definecolor{orange}{HTML}{C2570E}
\definecolor{bluet}{HTML}{294483}
\definecolor{green}{HTML}{298323}
\definecolor{red}{HTML}{A31313}
\setlength{\tabcolsep}{0.4em}
\captionsetup[table]{justification=raggedright,singlelinecheck=off}
\def\arraystretch{1.3}
\begin{table*}[ht!]

\begin{center}
  \caption{EfficientPhys and POS on transformed (deteriorated) UBFC-rPPG, UCLA-rPPG and UBFC-Phys}
  \label{tab:p_data2}
\begin{tabular}{c|ccc|ccc|ccc|ccc|ccc|}
    
    
    \cline{2-16}
      \multicolumn{1}{c|}{}
    & \multicolumn{3}{c|}{UBFC-rPPG~\cite{ubfc}}
    & \multicolumn{3}{c|}{UCLA-rPPG~\cite{ucla}} 
    & \multicolumn{3}{c|}{UBFC-Phys T1}
    & \multicolumn{3}{c|}{UBFC-Phys T2}
    & \multicolumn{3}{c|}{UBFC-Phys T3}
    \\                                             
    
    \cline{2-16}      
       \multicolumn{1}{c|}{} 
    & \multicolumn{1}{c|}{MAE} 
    & \multicolumn{1}{c|}{RMSE} 
    & \multicolumn{1}{c|}{MAPE}
    & \multicolumn{1}{c|}{MAE} 
    & \multicolumn{1}{c|}{RMSE} 
    & \multicolumn{1}{c|}{MAPE} 
    & \multicolumn{1}{c|}{MAE} 
    & \multicolumn{1}{c|}{RMSE} 
    & \multicolumn{1}{c|}{MAPE}
    & \multicolumn{1}{c|}{MAE} 
    & \multicolumn{1}{c|}{RMSE} 
    & \multicolumn{1}{c|}{MAPE}
    & \multicolumn{1}{c|}{MAE} 
    & \multicolumn{1}{c|}{RMSE} 
    & \multicolumn{1}{c|}{MAPE}
    \\ 

\hline
\multicolumn{16}{c}{\textcolor{green}{Original data}} \\ 
\hline
EfficientPhys~\cite{xiuliu2021} & \textbf{\uline{5.11}} & \textbf{\uline{13.05}} & \textbf{\uline{4.75}}  & \uline{0.57} 
                                & \uline{1.74} & \uline{0.81} & \uline{4.35} & \uline{9.46} & \uline{6.65} & \uline{15.32} 
                                & \uline{19.91} & \uline{23.99}  & 15.65 & \uline{20.38} & 23.04 \\
POS~\cite{wang2017} & 9.75 & 19.27 & 8.88 & 1.46 & 4.46 & 1.93  & 6.03 & 13.93 & 8.52 & 15.57 & 20.55 & 22.64 & \uline{14.65} 
                    & 20.42 & \uline{21.97} \\

\hline
\multicolumn{16}{c}{\textcolor{orange}{Gaussian blur data}} \\ 
\hline
EfficientPhys~\cite{xiuliu2021} & \uline{5.30} & \uline{13.18} & \uline{4.97} & \uline{\textbf{0.53}} & \uline{\textbf{1.61}} 
                                & \uline{\textbf{0.76}} & \uline{4.65} & \uline{9.99} & \uline{7.25} & \uline{14.15} & \uline{19.61} 
                                & \uline{22.37}  & 15.23 & 20.22 & 22.75 \\
POS~\cite{wang2017} & 10.22 & 20.32 & 9.32 & 1.48 & 4.57 & 2.05 & 6.03 & 13.93 & 8.52 & 16.11 & 20.97 & 23.58 & \uline{14.44} 
                    & \uline{20.29} & \uline{21.65} \\

\hline
\multicolumn{16}{c}{\textcolor{red}{Gaussian noise data}} \\ 
\hline
EfficientPhys~\cite{xiuliu2021} & 19.69 & 24.71 & 20.21 & 14.54 & 20.37 & 21.80 & 12.64 & 17.27 & 17.14 & 17.54 & 20.75 
                                & \uline{16.13} & 17.08 & 21.11 & 24.60 \\
POS~\cite{wang2017} & \uline{18.66} & \uline{26.62} & \uline{16.92} & \uline{4.4} & \uline{8.29} & \uline{6.02} & \uline{8.37} 
                    & \uline{15.27} & \uline{10.90} & \uline{14.69} & \uline{18.90} & 22.12 & \uline{14.40} & \uline{18.29} 
                    & \uline{21.37} \\

\hline
\multicolumn{16}{c}{\textcolor{blue}{Eyemask data}} \\ 
\hline
EfficientPhys~\cite{xiuliu2021} & \uline{5.39} & \uline{13.66} & \uline{5.01} & \uline{1.29} & \uline{3.74} & \uline{1.78} 
                                & \uline{5.39} & \uline{13.66} & \uline{5.01} & 15.40 & 20.46 & 22.28  & \uline{13.94} & \uline{20.24} 
                                & \uline{21.01} \\
POS~\cite{wang2017} & 12.38 & 21.63 & 11.61 & 1.46 & 4.07 & 1.89 & 12.38 & 21.63 & 11.61 & \uline{13.14} & \uline{17.70} 
                                & \uline{19.20} & 15.99 & 21.22 & 23.04 \\

\hline
\multicolumn{16}{c}{\textcolor{purple}{Facemask data}} \\ 
\hline
EfficientPhys~\cite{xiuliu2021} & \uline{10.69} & \uline{18.71} & \uline{10.56} & 13.13 & 23.91 & 18.83 & \textbf{\uline{4.31}} 
                                & \uline{\textbf{9.40}} & \textbf{\uline{6.62}} & 19.71 & 24.44 & 28.61 & 18.37 & 23.11 & 25.02 \\
POS~\cite{wang2017} & 15.38 & 25.28 & 13.86 & \uline{2.92} & \uline{6.74} & \uline{3.78} & 6.07 & 13.10 & 8.32 & \uline{\textbf{12.56}}                     & \uline{\textbf{17.10}} & \uline{\textbf{18.54}} & \uline{\textbf{13.02}} & \uline{\textbf{17.23}} 
                     & \uline{\textbf{18.84}} \\

\hline
\end{tabular}
\end{center}
\vspace{1mm}
\footnotesize{\textbf{Note:} The bold number represents the best result of five types of data, while the underlined number represents the best result between the methods.}
\vspace{-5mm}
\end{table*}

\subsection{rPPG Evaluation on Restored Datasets}
\label{sec:restored}

\begin{figure}
    \centering
     \begin{subfigure}[b]{\columnwidth}
         \centering
         \includegraphics[width=\columnwidth]{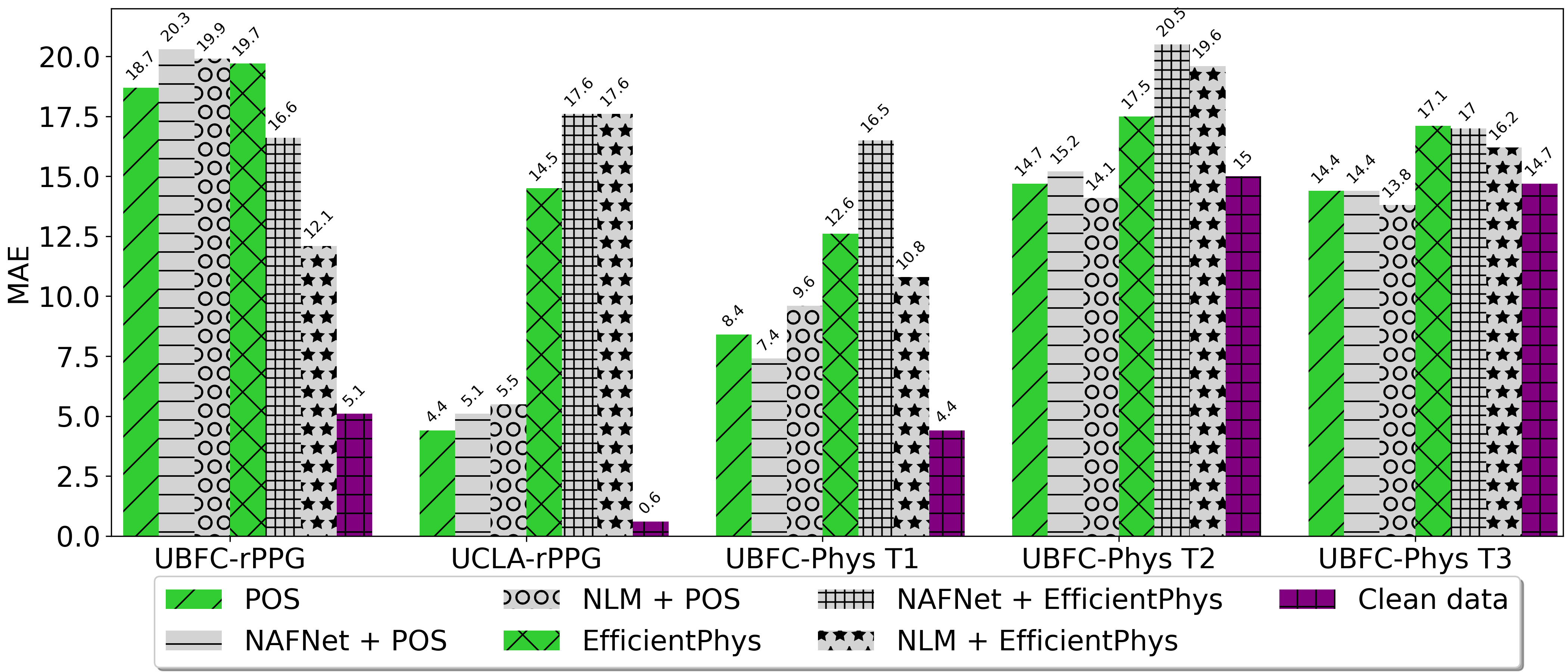}
         \caption{MAE}
         \label{fig:Gaussian_Restored_MAE}
     \end{subfigure}
     \begin{subfigure}[b]{\columnwidth}
         \centering
         \includegraphics[width=\columnwidth]{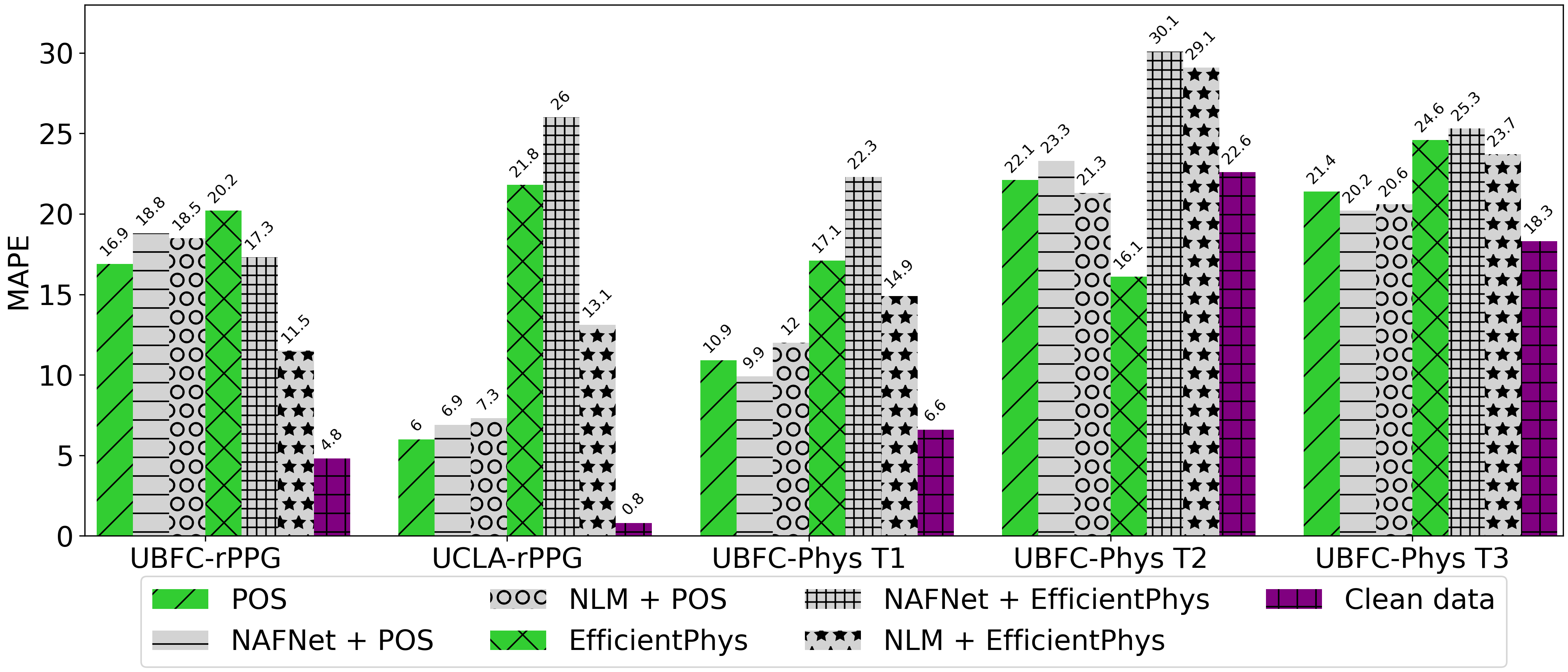}
         \caption{MAPE}
         \label{fig:Gaussian_Restored_MAPE}
     \end{subfigure}
     \begin{subfigure}[b]{\columnwidth}
         \centering
         \includegraphics[width=\columnwidth]{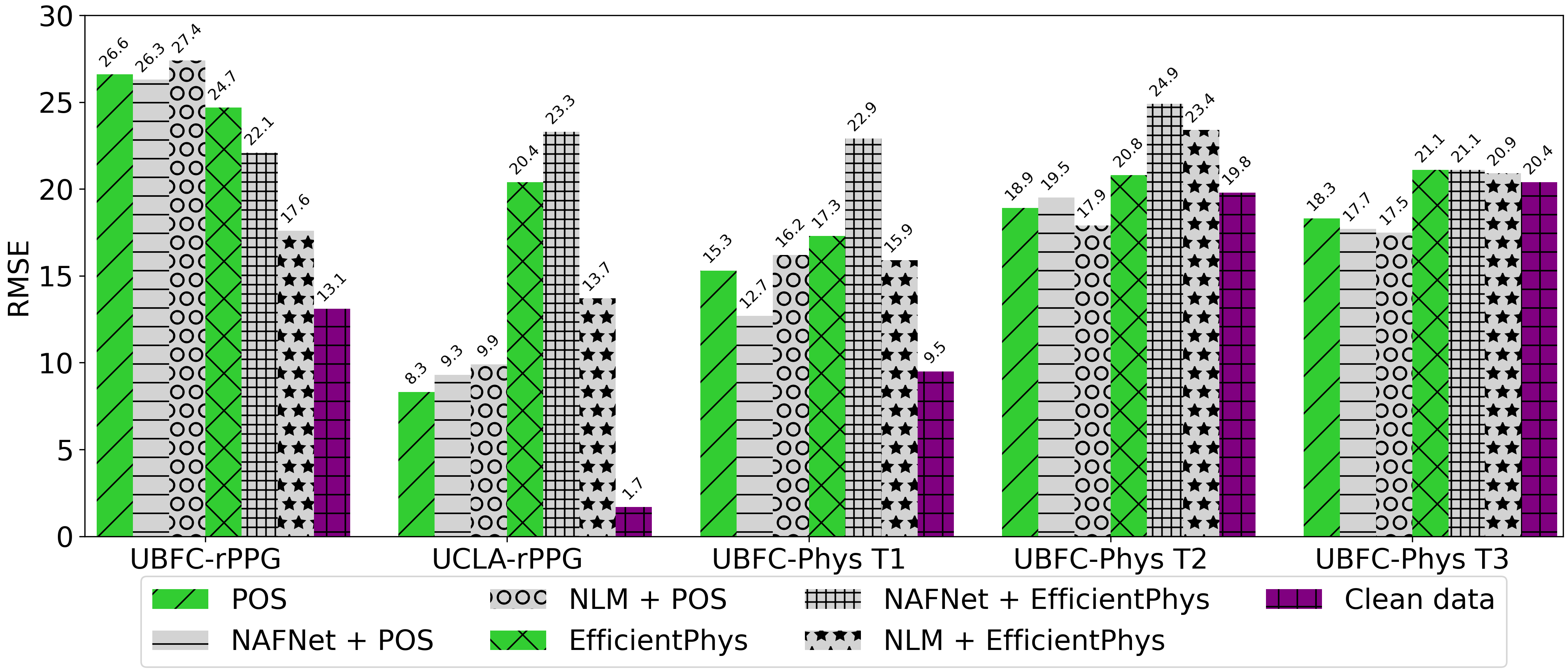}
         \caption{RMSE}
         \label{fig:Gaussian_Restored_RMSE}
     \end{subfigure}
    \caption{Gaussian noise data - results with/without restoration}
    \label{fig:Gaussian_Restored}
\end{figure}

\begin{figure}[t]
    \centering
     \begin{subfigure}[b]{\columnwidth}
         \centering
         \includegraphics[width=\columnwidth]{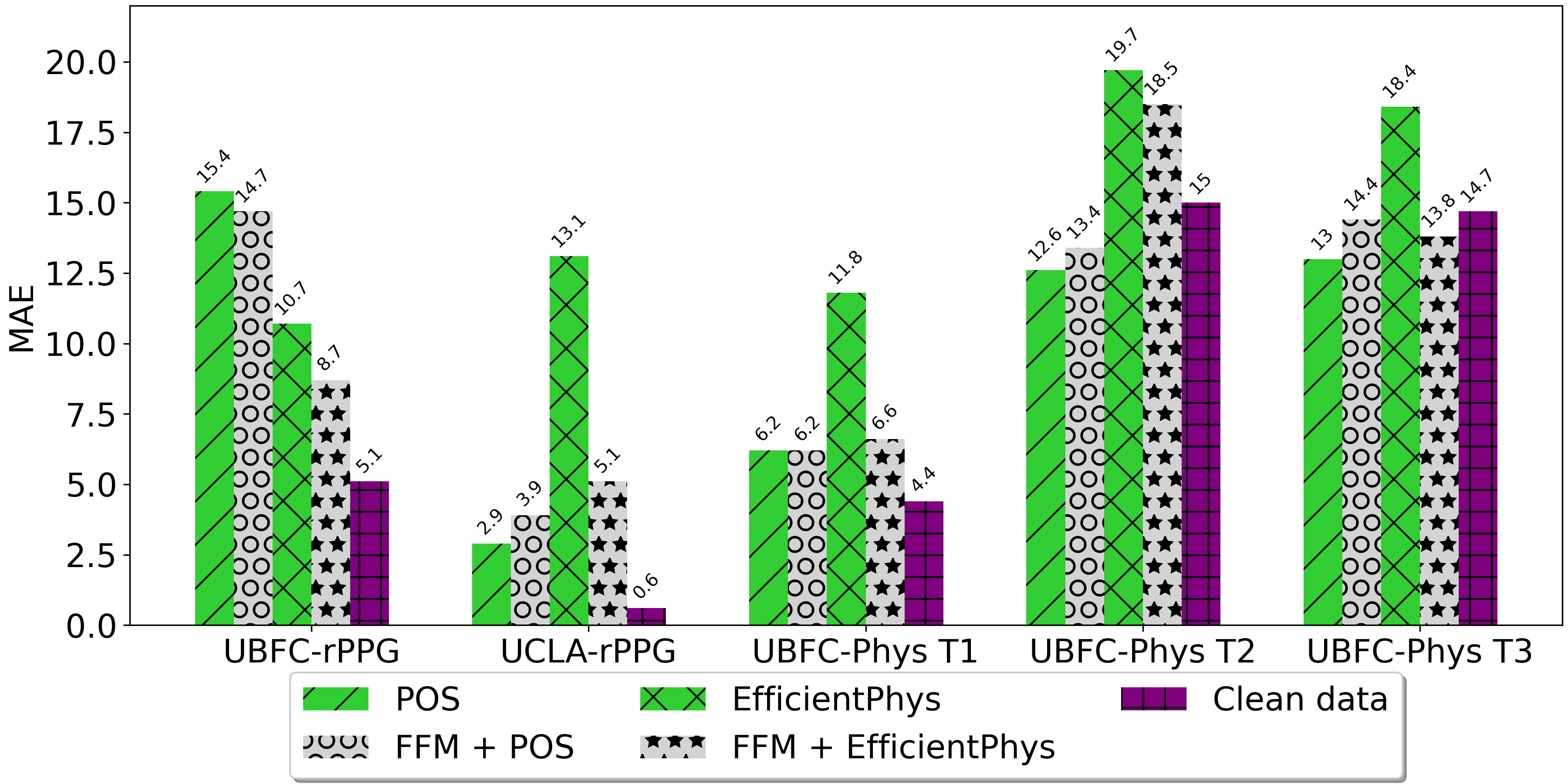}
         \caption{MAE}
         \label{fig:Facemask_Restored_MAE}
     \end{subfigure}
     \begin{subfigure}[b]{\columnwidth}
         \centering
         \includegraphics[width=\columnwidth]{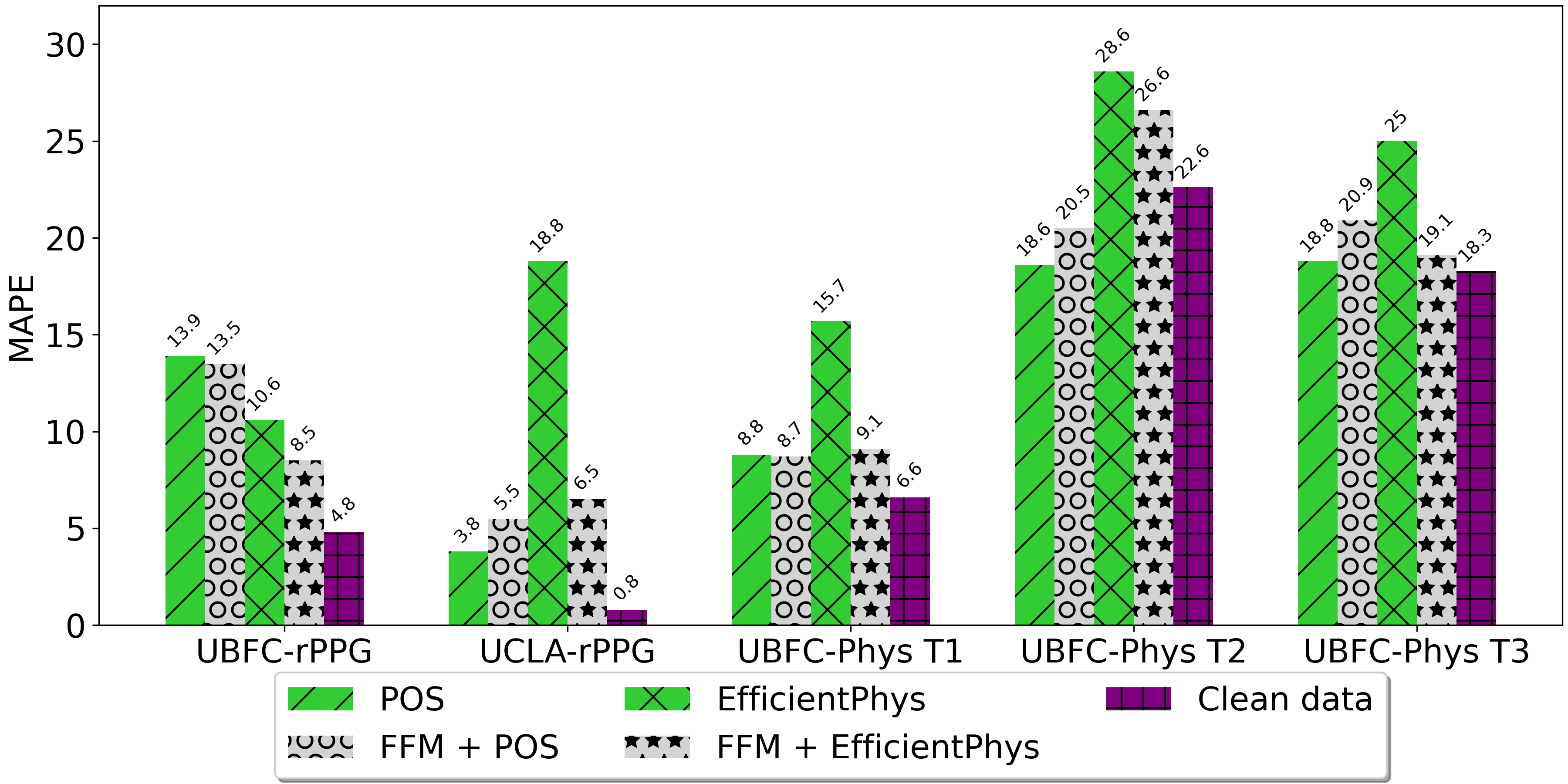}
         \caption{MAPE}
         \label{fig:Facemask_Restored_MAPE}
     \end{subfigure}
     \begin{subfigure}[b]{\columnwidth}
         \centering
         \includegraphics[width=\columnwidth]{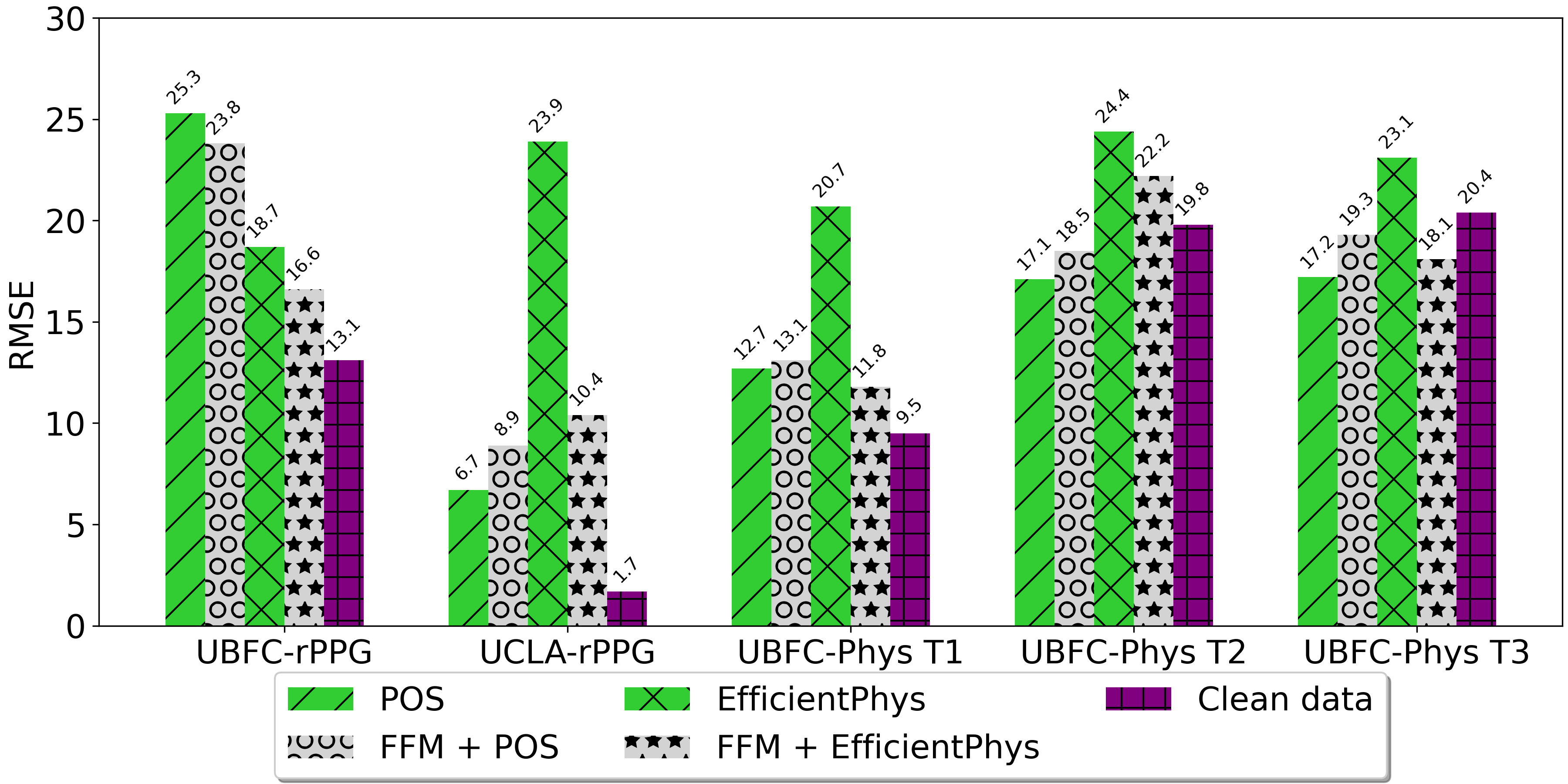}
         \caption{RMSE}
         \label{fig:Facemask_Restored_RMSE}
     \end{subfigure}
    \caption{Facemask data - results with/without restoration}
    \vspace{-7mm}
    \label{fig:Facemask_Restored}
\end{figure}

In Figure~\ref{fig:Gaussian_Restored}, we summarized the rPPG results on the videos deteriorated with Gaussian noise.
We added the performance of the original datasets (as clean data) from Table~\ref{tab:p_data1} for comparison.
The EfficientPhys pre-trained model performs well on restored datasets, with significant improvements, even if the denoised image is blurred.
It produced the best results on the UBFC-rPPG dataset.
However, its improvement is not significant for the UBFC-Phys T2 dataset.
In terms of the POS method, the NAFNet and NLM methods do not offer much improvement in the UBFC-rPPG and UCLA-rPPG datasets, and only a slight improvement is observed with the NAFNet method in the UBFC-Phys T1 dataset. Although the NLM method helps improve the performance of POS for datasets T2 and T3, the improvement is only limited.

In Figure~\ref{fig:Facemask_Restored}, FFM produces significant improvements in most datasets when combined with EfficientPhys. In particular, the UBFC-Phys T1 and T3 datasets show results that are close to, or even better than, those produced by the POS method in some metrics. The result on the UBFC-rPPG dataset also shows this combination is much better than using the POS method, while the UCLA-rPPG dataset shows an improvement of more than double, although still not quite as good as the POS method.
One reason for the superior of the POS method over the learning-based approach is that the POS method~\cite{wang2017} extract the pulse signal from skin regions; hence, the facemask does not effect its accuracy as significantly as the learning-based method.
On the other hand, the results indicates that combining the FFM and POS methods may not be optimal.


\section{Related work}
The robustness of image processing systems is an essential research topic. One direction is to design or learn invariant features that do not impair the performance of these systems in challenging scenarios. For example, Flusser~\textit{et al.}~\cite{Flusser2016} proposed Gaussian blur-invariant features for template-based image recognition. Hendrycks and Dietterich~\cite{Hendrycks2019} assessed the robustness of neural network classifiers in the presence of several image corruptions and perturbations, including noise, blur, weather, and digital categories. Corrupted images could be considered as data augmentation for training machine learning models. Mixing augmented and clean data allowed the models to learn robust 
features~\cite{Hendrycks2021}. These approaches focused on classification while ours was a regression setting.

Face covering accessories such as facemasks have caused difficulty for rPPG methods.
Speth~\textit{et al.}~\cite{Speth2022} implemented remote pulse estimation on videos of human faces with facemasks. Limitations of equipment and networking facilities also influence the accuracy of rPPG. Álvarez~\textit{et al.}~\cite{Alvarez2023} studied remote heart-rate measurement in the presence of frame-dropping and low resolution. Our work takes into account several transformations that cover the faces, such as facemasks and sunglasses and different effects such as noise or blurring. 

Privacy is another issue to consider when processing face images. Abbasi~\textit{et al.}~\cite{Abbasi2022} blurred photos at various degrees, (equivalent to the privacy levels) to protect sensitive facial images. They formulated trade-off criteria to optimize the privacy and accuracy of face recognition frameworks. Our study considers not only blurring but also noise addition and face-covering cases such as facemasks and sun-glasses.

\section{Conclusion}
In this study, we utilized image processing techniques to intentionally degrade video quality to test the performance of rPPG non-learning methods and pre-trained models under challenging conditions. We then implemented image restoration methods such as noise reduction and facemask removal to improve the accuracy of heart-rate estimation in deteriorated videos. In most of the datasets, our findings indicate that combining the rPPG methods (POS or EfficientPhys) and the FFM restoration technique produced the most accurate results on degraded face videos. In several cases, the non-learning POS method achieved better results since it only focuses on skin regions and excludes facemasks. The success of our approach provides valuable insights into the potential of advanced image restoration techniques to improve the performance of non-contact heart rate monitoring methods in real-world scenarios. Furthermore, our results demonstrate the feasibility of developing privacy-aware rPPG techniques that can be applied to face-covering videos.



\bibliographystyle{IEEEtran}
\bibliography{citations.bib}
\end{document}